\documentclass[letterpaper]{article} 
\usepackage{aaai2026}  
\usepackage{times}  
\usepackage{helvet}  
\usepackage{courier}  
\usepackage[hyphens]{url}  
\usepackage{graphicx} 
\usepackage{natbib}  
\usepackage{caption} 
\usepackage{algorithm}
\usepackage{algorithmic}

\usepackage{newfloat}
\usepackage{amsmath} 
\usepackage{pifont}
\usepackage{multirow}
\usepackage{threeparttable}
\usepackage{booktabs}
\usepackage{amssymb}
\usepackage{listings}
\usepackage{xcolor}
\usepackage{makecell}

\usepackage{newfloat}
\usepackage{listings}
\DeclareCaptionStyle{ruled}{labelfont=normalfont,labelsep=colon,strut=off} 
\floatstyle{ruled}
\newfloat{listing}{tb}{lst}{}
\floatname{listing}{Listing}
\title{AAAI Press Formatting Instructions \\for Authors Using \LaTeX{} --- A Guide}
\author{
    \textbf{Lichen Ma}\textsuperscript{\rm 1}\equalcontrib,
    \textbf{Xiaolong Fu}\textsuperscript{\rm 1}\equalcontrib,
    \textbf{GaojingZhou}\textsuperscript{\rm 1},
    \textbf{Zipeng Guo}\textsuperscript{\rm 1,2},
    \textbf{Ting Zhu}\textsuperscript{\rm 1},
    \textbf{Yichun Liu}\textsuperscript{\rm 1},\\
    \textbf{Yu Shi}\textsuperscript{\rm 1},
    \textbf{Jason Li}\textsuperscript{\rm 1},
    \textbf{Junshi Huang}\textsuperscript{\rm 1}\thanks{Corresponding Author.}
}

\affiliations{
    \textsuperscript{\rm 1}JD.COM \\ 
    \textsuperscript{\rm 2}Sun Yat-sen University \\
    \{malichen2020, fxlcumt, junshi.huang\}@gmail.com
}

\DeclareCaptionStyle{ruled}{labelfont=normalfont,labelsep=colon,strut=off} 
\floatstyle{ruled}
\newfloat{listing}{tb}{lst}{}
\floatname{listing}{Listing}

\title{UM-Text: A Unified Multimodal Model for Image  Understanding \\ and Visual Text Editing}

\usepackage{bibentry}
\twocolumn

\begin{document}

\maketitle

\begin{abstract}

With the rapid advancement of image generation, visual text editing using natural language instructions has received increasing attention. 
The main challenge of this task is to fully understand the instruction and reference image, and thus generate visual text that is style-consistent with the image. 
Previous methods often involve complex steps of specifying the text content and attributes, such as font size, color, and layout, without considering the stylistic consistency with the reference image.
To address this, we propose UM-Text, a \textbf{u}nified \textbf{m}ultimodal model for context understanding and visual text editing by natural language instructions. 
Specifically, we introduce a Visual Language Model (VLM) to process the instruction and reference image, so that the text content and layout can be elaborately designed according to the context information.
To generate an accurate and harmonious visual text image, we further propose the UM-Encoder to combine the embeddings of various condition information, where the combination is automatically configured by VLM according to the input instruction.
During training, we propose a regional consistency loss to offer more effective supervision for glyph generation on both latent and RGB space, and design a tailored three-stage training strategy to further enhance model performance. 
In addition, we contribute the UM-DATA-200K, a large-scale visual text image dataset on diverse scenes for model training.
Extensive qualitative and quantitative results on multiple public benchmarks demonstrate that our method achieves state-of-the-art performance.
\end{abstract}

\section{Introduction}
\begin{figure}[h]
    \centering
    \includegraphics[width=0.47\textwidth]{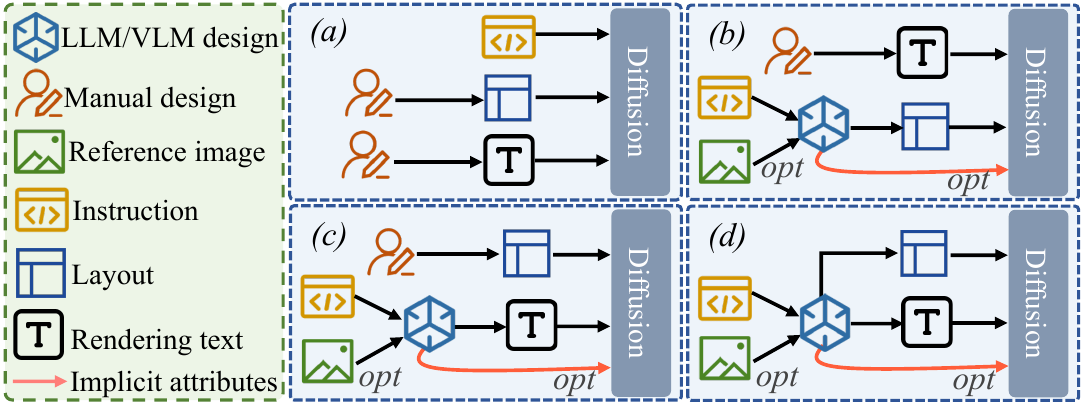}
    \caption{Illustration of traditional framework of visual text generation and three additional generation patterns of our method. Please note that the text content, layout and implicit attributes can be adaptively generated by VLM according to instruction.}
    \label{fig:model_input}
\end{figure}
Visual text editing and generation play a crucial role in various applications, such as poster design, scene text editing, and the novel task of cross-language image translation. 
The main challenge of these tasks lies in manual design of text layout, attributes (\textit{e.g.}, font type, size, color), language (\textit{e.g.}, English, Chinese), and visual context (\textit{e.g.}, poster, product image), which are cumbersome and error-prone. 
In this paper, we propose a method that enables users to perform visual text editing via natural language instructions. 
Given an input image and editing command, our model automatically generates compelling text content, appropriate layout, and visually harmonious text images with implicit text attributes.

\begin{figure*}[h]
    \centering
    \includegraphics[width=0.91\textwidth]{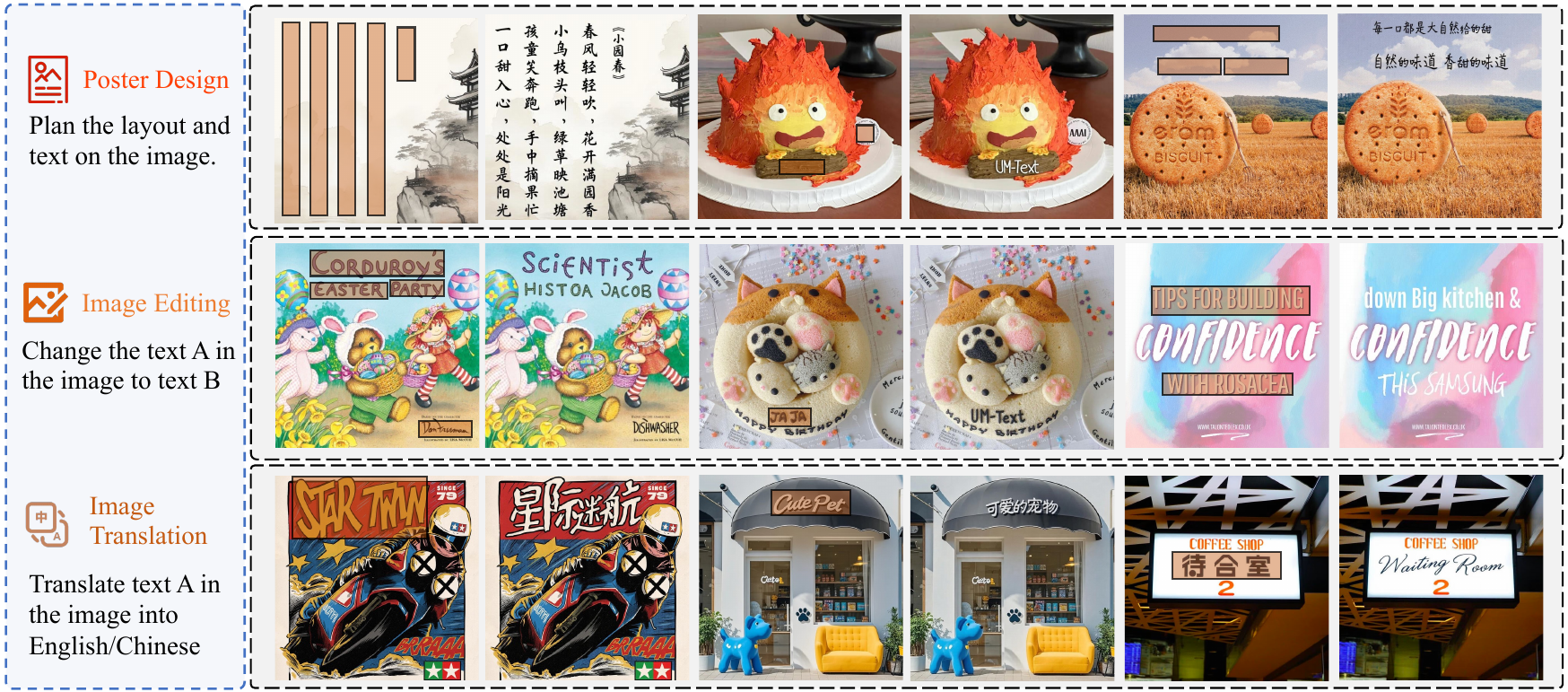}
    \caption{Some results produced by our UM-Text, presenting its powerful effects on tasks such as image editing, image translation, and poster design. 
    Please note that the bounding boxes of text are adaptively generated by UM-Text model.}
    \label{fig:teaser}
\end{figure*}

Recently, with the rapid advancement of text-to-image (T2I), diffusion models have enabled the creation of highly realistic images with only instructions. 
For example, Stable Diffusion 3\cite{esser2024scaling}, FLUX.1\cite{flux.1} and FLUX.1 Kontext\cite{labs2025fluxkontext} has gradually improved its capabilities in general image generation and visual text rendering.
However, these commonly used T2I models are still deficient in generating complex characters such as handwriting or art text.
Many researchers inject enhanced text generation capabilities into pre-trained diffusion models using various approaches~\cite{tuo2023anytext,tuo2024anytext2,chen2023textdiffuser,chen2024textdiffuser}, but manual interactions are still required for text content and layout.  

In visual text generation, text layout plays an important role in generating an appropriate result. 
Some text generation methods, such as TextDiffuser2\cite{chen2024textdiffuser}, UniGlyph\cite{wang2025uniglyph}, and GlyphDraw2\cite{ma2025glyphdraw2}, use the task-specific large language model (LLM) to predefine text positions.
DesignDiffusion\cite{wang2025designdiffusion} even achieves the layout and visual text image in an end-to-end manner.
However, there approaches are still infeasible for visual text editing task, which requires the coordinates of the target text in image.
Moreover, the potential of text editing task should be further explored to generate more harmonious and aesthetic visual text images.

In this paper, we propose a holistic framework that integrates multimodal understanding into the process of visual text generation and editing for implicit learning of text layout and attributes.
As illustrated in Fig.\ref{fig:model_input}, our framework can support four different text generation and editing patterns, where the reference image and instruction information is extracted as context embeddings for visual text generation/editing.
Furthermore, we introduce UM-Encoder, a module for multiple condition aggregation that incorporates T5 embeddings, character-level visual embeddings, and context embeddings.
To improve the accuracy of text glyphs, the regional consistency loss in both latent and RGB space are proposed for better glyph supervision.
We also contribute the UM-DATA-200K dataset containing 200k diverse image pairs with/without visual text for the pre-training of VLM. 

In summary, our contributions are as follows.

\begin{itemize}
\item We propose an innovative framework named UM-Text, which combines a unified multimodal understanding and image editing model. With a three-stage training strategy and region-based losses, UM-Text allows flexible visual text generation and editing by simple natural language instructions.

\item We introduce UM-Encoder, a novel module for multiple condition aggregation that integrates text embedding, character-level visual embeddings, and multimodal embeddings.
With this module, the implicit attributes and layout of text are adaptively generated for visual text generation and editing.

\item We contribute a dataset called UM-DATA-200K with manual annotation for visual text generation and editing. Extensive experiments demonstrate the effectiveness of our dataset and framework.
\end{itemize}

\begin{figure*}[h]
    \centering
    \includegraphics[width=0.99\textwidth]{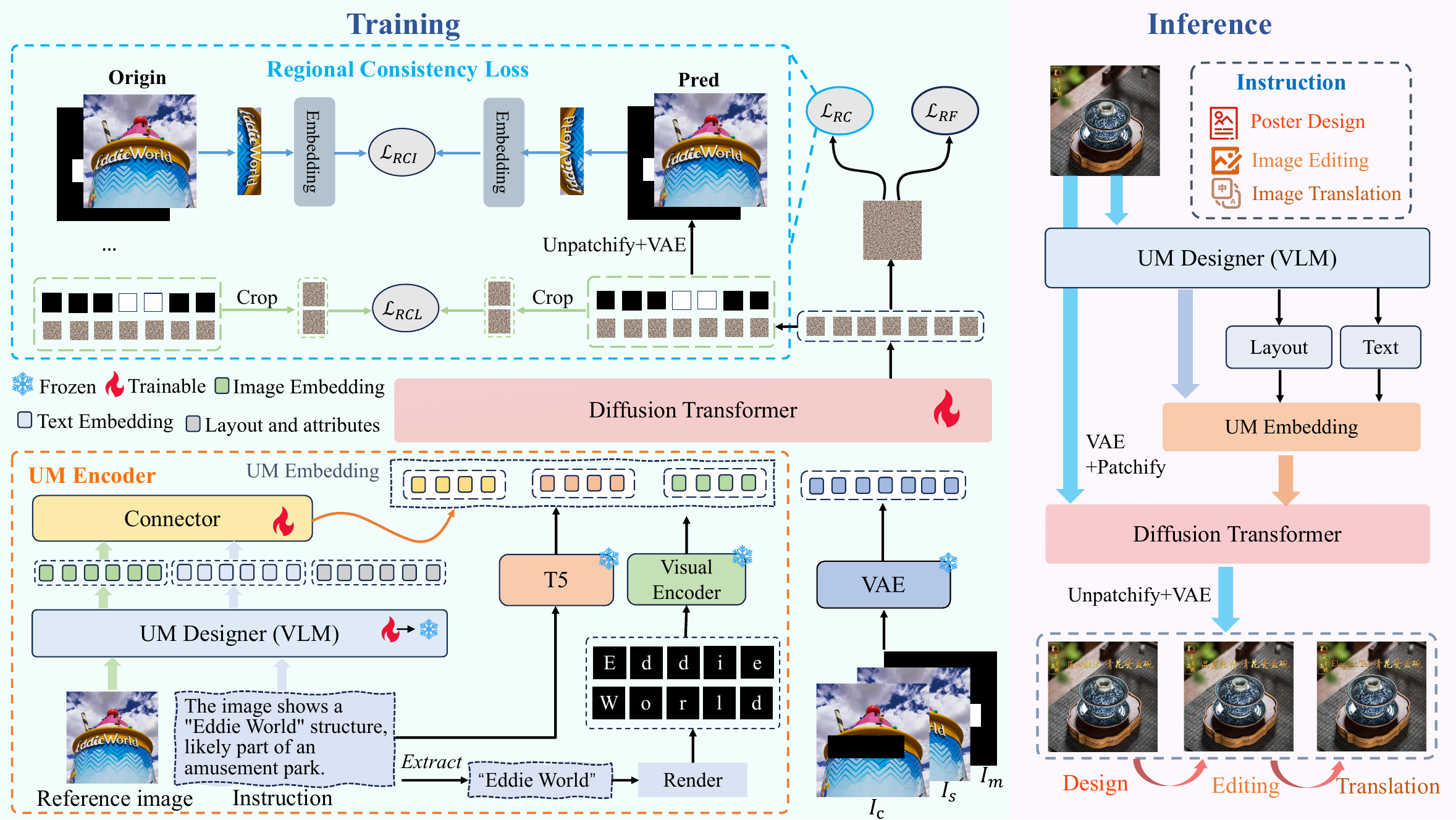 }
    \caption{The framework of UM-Text for multi-lingual visual text generation and editing.
    The UM-Encoder integrates multiple modality embeddings as the condition of visual text generation.
    The mask in input and loss function is transformed from the predicted layout of UM-Designer.
    Please note our single model supports diverse downstream applications based on the instructions.}
    \label{fig:main}
\end{figure*}

\section{Related Work}
\subsubsection{Image Generation and Understanding}

Diffusion models have become the primary method for high-quality image synthesis, offering powerful capabilities in terms of photorealism, fidelity, and diversity. From DDPM \cite{ho2020denoising} and DDIM \cite{song2020denoising} to Latent Diffusion Models (LDM) \cite{podell2023sdxl,rombach2022high,tian2024visual}, these models improve generation efficiency and scalability by operating directly within the latent embedding space, enabling image synthesis with higher resolution at lower computational costs. With the introduction of architectures such as DiT \cite{peebles2023scalable,esser2024scaling} and FLUX \cite{flux.1}, diffusion models have made significant advances in generalization and image quality, laying a solid foundation for unified handling of multimodal conditions and becoming an important architecture in modern image generation \cite{labs2025flux, t2i,zhang2023adding}. Despite these advancements, diffusion models still face significant challenges in understanding textual and visual information, highlighting the need to introduce VLM.

VLM \cite{bai2025qwen2.5vl,team2025kimi,zhu2504internvl3,gpt4o,team2023gemini} have made significant progress in vision-language understanding tasks. Models like Gemini \cite{team2024gemini}, Janus-Pro \cite{chen2025janus}, Mogao \cite{liao2025mogao}, BAGEL \cite{zhang2025unified} and Nexus-Gen \cite{zhang2025nexus} further unify understanding and generation. Recent works such as MetaQueries \cite{MetaQueries}, BLIP3o \cite{chen2025blip3}, UniWorld-V1 \cite{2025UniWorld}, OmniGen2 \cite{wu2025omnigen2}, and Step1X-Edit \cite{liu2025step1x-edit} integrate VLMs into image generation via multimodal conditioning, exploring control mechanisms and latent-level fusion with diffusion models. However, these methods still face limitations in text rendering: most of them only support English and struggle with editing fine-grained textual regions.

\subsubsection{Visual Text Generation and Editing}
In recent years, there have been substantial developments in the tasks of T2I generation and image editing with visual text rendering. The goal of visual text generation and editing models is to produce accurate text images where the visual elements and text layout are harmoniously integrated. Text embeddings and various loss functions are employed to help the model generate more precise text. 

DrawText\cite{liu2022character} demonstrates that character-aware models consistently outperform their character-blind counterparts across various text rendering tasks. Glyph-ByT5\cite{liu2024glyph,liu2024glyph2} and FLUX-Text\cite{lan2025flux} introduce a method utilizing box-level contrastive learning to align text features extracted from the language model with those derived from the visual encoder. In DiffUTE\cite{chen2023diffute} and GlyphDraw\cite{ma2023glyphdraw}, glyph images are directly incorporated into the text embeddings. AnyText\cite{tuo2023anytext}, AnyText2\cite{tuo2024anytext2}, and GlyphDraw2\cite{ma2025glyphdraw2} render a glyph line containing multiple characters into an image, encode glyph information using a pretrained OCR recognition model, and inject it into the text embedding. 

Unfortunately, it’s challenging to represent multiple characters with a single token, and there’s a lack of embedding for image content. To address these issues, we propose UM-Encoder for text embedding injection that integrates T5 embeddings, character-level visual embeddings, and VLM embeddings. This approach enables the model to generate more accurate text while achieving better stylistic consistency with the reference image. 

Several T2I methods employ large language models (LLMs) for layout prediction. TextDiffuser has adopted a Layout Transformer that autoregressively outputs bounding boxes for keywords in an encoder-decoder manner. GlyphDraw2 and TextDiffuser2 further leverage LLMs to generate layouts. However, these methods simply learn layout information from the text modality and cannot be directly applied to visual text editing tasks. To overcome this limitation, we propose UM-Designer, a VLM that can simultaneously generate layouts and text related to the reference image.

\begin{figure*}[t]
    \centering
    \includegraphics[width=1\textwidth]{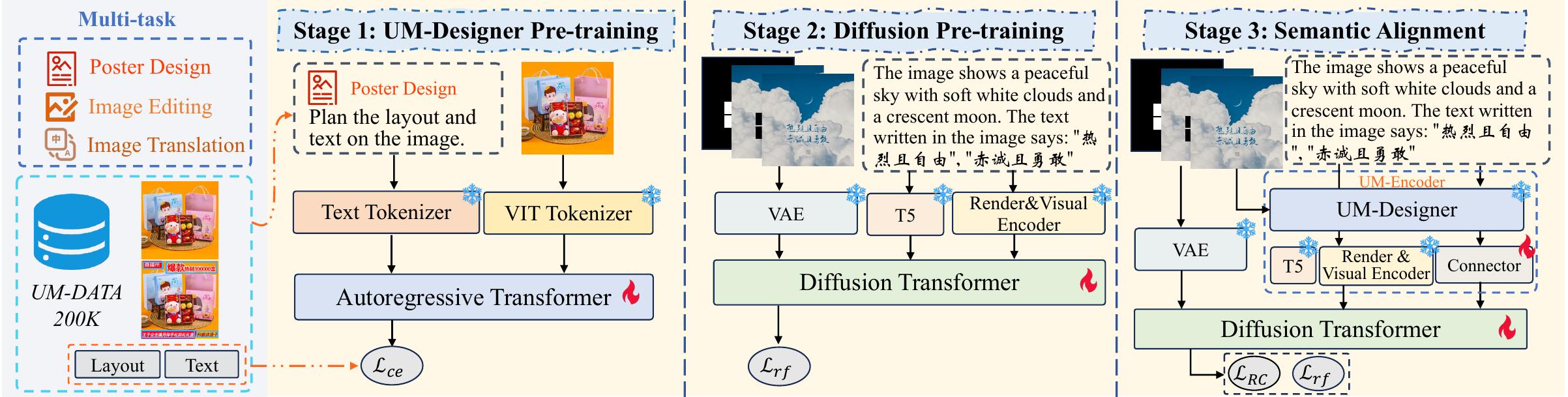}
    \caption{The illustration of three training stages for UM-Text optimization.}
    \label{fig:train_3_stage}
\end{figure*}

\section{Methodology}
In this section, we present the details of UM-Text.
We begin to introduce the construction process of UM-DATA-200K in Sec.\ref{Sec.3.1}, which is a large-scale synthetic dataset designed to pretrain the UM-Designer with capabilities in layout planning and text content generation.
In Sec.\ref{Sec.3.2}, we present the framework of our UM-Text for visual text generation and editing tasks. 
Subsequently, Sec.\ref{Sec.3.3} introduces the UM-Encoder, which integrates various conditions into unified embeddings.
In Sec.\ref{Sec.3.4}, we propose a region-wise consistency loss to ensure that the generated text is semantically accurate and stylistically consistent with the reference image.
Finally, Sec.\ref{Sec.3.5} outlines our training strategy.

\subsection{Dataset Construction}\label{Sec.3.1}
Recently, many layout planning and text content generation datasets are limited in the scale or quality of collected data.
To address this gap, this paper endeavors to assemble a large-scale, high-quality dataset particularly tailored for layout planning, visual text generation, and editing tasks.
Generally, we crawled 40 million product posters from online e-commerce platforms. 
To construct our dataset, we developed an advanced data pipeline including image aesthetics filtering, object segmentation, OCR, image erasure, and manual annotation.

Specifically, we used the PPOCRv4\cite{cui2025paddleocr30technicalreport} to extract text content and bounding boxes from images and employed Aesthetic Predictor V2.5 to rate the images.
We utilized OCR results and aesthetic scores to filter five million images with detailed text layouts and contents.
To achieve higher-quality images, we further applied SAM2\cite{2024SAM} to segment the main product, filtered out inconsistent text layouts, and used FLUX-Fill \cite{flux.1-fill} to generate clean images based on the text layout.
Ultimately, we selected 200k images, including various styles of main product and poster images.

\subsection{The Framework of UM-Text }\label{Sec.3.2}
As illustrated in Fig.~\ref{fig:main}, the main components of UM-Text include the UM-Encoder, the Diffusion Transformer, and training losses for optimization.
Generally, UM-Designer is implemented as a VLM to capture the semantic information of instruction and reference image for the prediction of text content, layout, and implicit attributes.
In addition to the instruction embedding from T5 and visual embedding of rendering text images, these predicted results are adaptively selected as additional conditions for downstream tasks according to the instruction, all of which constitute the conditions of diffusion model, named UM-Embedding $c_e$.

In the flow-matching-based diffusion model, we use VAE encoder to extract the latent representations of input image $I_s$, binary mask image $I_m$ from UM-Designer or manually designed layout, and condition image $I_c = I_s \odot I_m$, resulting in $z_0$, $z_m$ and $z_c$, respectively.
Subsequently, the diffusion algorithm progressively adds random noise to $z_0$ at each time step $t$, resulting in a series of noisy latent variables $z_t$.
Flow-based diffusion models employ a neural network $V_\theta$ to predict the velocity field at each time step, with the objective of matching the model’s velocity field to the ideal velocity field that transports the data distribution along the diffusion process.
This is achieved by minimizing the flow matching loss:
\begin{equation}
\begin{split}
    \mathcal L_{RF} =  \mathbb{E}_{z_t,z_m,z_c,c_e,t \sim \mathcal N(0,1)} [||V^*(z_t,t)\\- V_\theta(z_t,z_m,z_c,c_e,t)||_2^2].
\end{split}
\end{equation}
where $\mathcal{L}_{RF}$ denotes the calibrated flow matching loss, and $V^*(z_t, t)$ is the target velocity field derived from the diffusion process.

\begin{figure*}[h]
    \centering
    \includegraphics[width=0.99\textwidth]{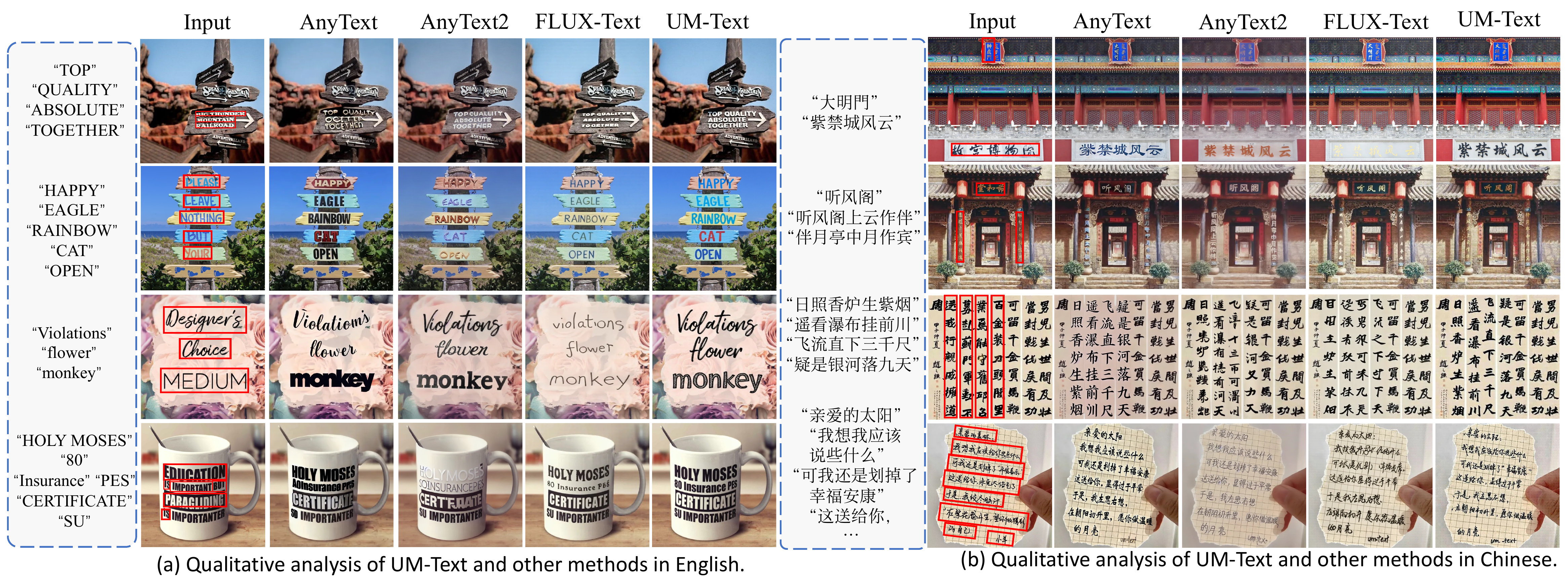}
    \caption{Qualitative comparison of UM-Text and state-of-the-art models in visual text editing task.}
    \label{fig:comppare_1}
\end{figure*}

\subsection{UM-Encoder}\label{Sec.3.3}
Currently, many ControlNet-like approaches\cite{ma2024chargen,wang2025dreamtext,zhao2023udifftext,chen2023diffute} typically inject glyph image and text conditions into the model. 
However, the glyph image condition is highly susceptible to the pre-defined text attributes, which often harm its robustness.
Some methods replace the text embedding with line-level OCR embedding as text condition. 
However, these approaches have some limitations:
(1) The visual embedding from OCR model only encodes the visual text information, missing the detailed description of generated image.
(2) Line-level visual embedding is insufficient for the representation of character stroke information.
(3) The layout and attributes of visual text are designed without considering the context information of reference image.

To address these issues, we propose the UM-Encoder for comprehensive condition representation on instruction, reference image, and character-level glyph image.
Specifically, we use a pretrained VLM, known as UM-Designer, to capture the semantic information of instruction and reference image.
The side information, including text content, layout, and attribute embeddings, for visual text generation/editing task is predicted by UM-Designer.
To obtain fine-grained glyph information of text, we render the text content into glyph images in character-level, and use an OCR model to extract the visual embeddings of glyph images.
Meanwhile, the output tokens of instruction and reference image are used as implicit attribute embeddings.
We claim that those token embeddings are effective enough for implicit representation of text attributes due to the well-designed pre-training task of UM-Designer. 
More details can be found in Sec~\ref{Sec.3.5}.
Finally, the character-level visual embeddings, attribute embeddings, and instruction embeddings from T5 are aligned and concatenated into UM-Embedding as the condition embeddings of diffusion model. 

\subsection{Regional Consistency Loss}\label{Sec.3.4}
Previous text generation methods often face challenges in generating correct strokes for complex characters, due to the lack of detailed supervision in nuanced glyph shapes.
To address this problem, we propose a Regional Consistency Loss (RC Loss) to constraint the structural consistency of visual text in various spaces.
Specifically, RC Loss receives the mask image $I_m$ from either the UM-Designer prediction result or manual annotation to localize the target regions, and calculate the $L_2$ distance between prediction result and ground-truth within target regions.

In our implementation, we design two types of RC Loss to constraint the structural consistency in both latent and RGB spaces.
In the latent space, we calculate the RC Loss in the velocity field, which is analogous to the re-weighting strategy of flow matching loss.
After that, we use VAE decoder to obtain the predicted image and use Canny edge detector to extract the edge maps of predicted image and input image.
Therefore, the RC Loss can be simply calculated on the localized regions of both edge images.
Formally, the RC Loss on latent and RGB spaces, denoted $\mathcal{L}_{RCL}$ and $\mathcal{L}_{RCI}$ respectively, can be formulated as:

\begin{equation}
\begin{split}
\mathcal{L}_{{RC}} = 
 \left\| \mathcal{C}(\hat{I} \odot I_m) - \mathcal{C}(I_s \odot I_m) \right\|_2^2 + \lambda \\ \mathbb{E} \left[ \left\| V^*(z_t, t)  \odot z_m 
- V_\theta(z_t, z_m, z_c, c_e, t) \odot z_m \right\|_2^2 \right]
\end{split}
\label{eq:regional_loss}
\end{equation} 
where $\hat{I}$ is the predicted image, and $\mathcal{C}(\cdot)$ denotes the Canny edge operator.
The overall training loss is defined as follows:
\begin{equation}
\mathcal L =  \mathcal L_{RF} + \beta \mathcal L_{RC}.
\end{equation}
and $\lambda$, $\beta$ are the hyper-parameters for balancing different losses.
This dual-space regional consistency loss effectively preserves stroke integrity in complex character generation while maintaining stability in the editing process.
Notably, the RC Loss in latent space mitigates the ``dilution effect" commonly observed in mask-based editing, where the gradients outside the mask dominate the optimization direction. 
\begin{table*} [h]
    
    \centering
    \begin{threeparttable}
    \begin{tabular*}{0.93\linewidth}{l|c|c|c|c|c|c|c|c|c}
    \toprule [1.5pt]
    \multirow{2}*{Methods}  & \multirow{2}*{Task} & \multicolumn{4}{c|}{English} & \multicolumn{4}{c}{Chinese} \\
    \cline{3-6} \cline{7-10}
    & & Sen.ACC $\uparrow$ & NED $\uparrow$ & FID $\downarrow$ & LPIPS $\downarrow$ & Sen.ACC $\uparrow$ & NED $\uparrow$ & FID $\downarrow$ & LPIPS $\downarrow$\\
    \midrule [1.5pt]
    GlyphControl & \multirow{3}*{T2I} & 0.5262 & 0.7529 & 43.10 & - & 0.0454 & 0.1017 & 49.51 & -  \\
    AnyText &   & 0.7239 & 0.8760 & 33.54 & - & 0.6923 & 0.8396 & 31.58 & -  \\
    AnyText-2 &   & 0.8096 & 0.9184 & 33.32 & - & 0.7130  & 0.8516 & 27.94 & -  \\
    \midrule [1.5pt]
    FLUX-Fill & & 0.3093 & 0.4698 & 33.87 & 0.1582 & 0.0292 & 0.0625 & 29.93 & 0.1207  \\
    AnyText & & 0.6843 & 0.8588 & 21.59 & 0.1106 & 0.6476  & 0.8210 & 20.01 & 0.0943  \\
    AnyText-2 & Editing & 0.7915 & 0.9100 & 29.76 & 0.1734 & 0.7022 & 0.8420 & 26.52 & 0.1444  \\
    FLUX-Text & & 0.8175 & 0.9193 & 12.35 & 0.0674 & 0.7213 & 0.8555 & 12.41 & 0.0487  \\
    UM-Text & & \textbf{0.8553} & \textbf{0.9395} & \textbf{10.15} & \textbf{0.0656} & \textbf{0.7988} & \textbf{0.8866} & \textbf{10.50} & \textbf{0.0481}  \\
    \bottomrule [1.5pt]
    \end{tabular*}

    \end{threeparttable}
    \caption{Comparison on AnyText-benchmark dataset.}\label{tab:anytext}
\end{table*}

\subsection{Training Strategy}\label{Sec.3.5}
Based on the model architecture, we propose a progressive three-stage training strategy to learn a text editing model with context-aware designing capabilities.
The training process is illustrated in Fig.\ref{fig:train_3_stage}, including the pre-training of UM-Designer, pre-training of diffusion model, and semantic aligment between UM-Designer and diffusion model.
The details are specified as follows.

Stage 1: UM-Designer Pre-training. 
In this stage, we initialize our UM-Designer by the weights of Qwen2.5-VL and continue the training process on UM-DATA-200K dataset.
Generally, this dataset contains various tasks, including layout planning, text content generation, text detection and recognition.
These tasks simulate the process of visual text generation and editing, and thus enhance the capability of UM-Designer for image-text understanding.

Stage 2: Diffusion Pre-training.
In this stage, we initialize the text generation model with FLUX-Fill and train all parameters on public benchmark.
This process enhances the foundational text generation capabilities for subsequent learning stage.

Stage 3: Semantic Alignment. 
In this stage, we train the connector of UM-Encoder and diffusion model to establish the connection between condition representation and application.
By further introducing VLM embedding, our UM-Embedding complements the vision-language understanding in large-scale text generation task, thereby enhances the glyph consistency and aesthetics of generated image.

Overall, with the structural vision-language guidance, detailed visual condition, and a powerful training strategy on unified framework, our UM-Text significantly improves the capability of model to generate high-fidelity and harmonious images in text editing task.

\section{Experiments}
\subsection{Implementation Details}
In stage 1 of training process, we initialize the weights of UM-Designer with Qwen2.5-VL-3B and train it on the UM-DATA-200K dataset for 10 epochs using 16 Tesla A100 GPUs.
In stage 2-3, we initialize the weights of the diffusion model with FLUX-Fill and train the model on AnyWord-3M dataset for 25 epochs.
After that, we introduce the UM-Designer of stage 1 into our framework, and train the connector and diffusion model on AnyWord-3M dataset for 5 epochs using 16 Tesla A100 GPUs.
The resolution of image is 512×512, and the resolution of rendering image for single character is 80×80. 
The strength coefficients $\lambda$ and $\beta$ are set to 5 and 2 by grid-search strategy, respectively.

\subsection{Dataset and Evaluation Metrics}
We use UMT-DATA-200K to train the UM-Designer model for layout and text design, and train UM-Text for visual text generation on AnyWord-3M \cite{tuo2023anytext}, which combines Wukong \cite{gu2022wukong}, LAION \cite{schuhmann2021laion}, and OCR-specific datasets (3M images). To ensure a fair comparison, UMT-DATA-200K is \textbf{not used} for visual text generation training in our experiments.

We evaluate on several public benchmarks following prior work. AnyWord-Benchmark \cite{tuo2023anytext} includes 1,000 English and 1,000 Chinese images. TextSeg \cite{xu2021rethinking} and LAION-OCR \cite{chen2023textdiffuser} provide 1,024 and 9.1M real-world text images, respectively. ICDAR13 \cite{karatzas2013icdar} contributes 233 test images for text detection evaluation. Following DREAMTEXT, we randomly select 100 images from the test sets of TextSeg, LAION-OCR, and ICDAR13 for evaluation.

AnyWord-Benchmark includes three evaluation metrics: Sentence Accuracy (Sen.ACC), Normalized Edit Distance (NED), and Frechet Inception Distance (FID) for distribution-level style similarity. We use Learned Perceptual Image Patch Similarity (LPIPS) to assess the consistency and realism of generated images, ensuring style consistency in edited regions while preserving non-target areas. All settings are consistent with FLUX-Text. 

Following DREAMTEXT, we use an off-the-shelf scene text recognition (STR) model to identify the rendered text and then evaluate word-level correctness using sequence accuracy (SeqAcc) by comparing the STR result with the ground truth.

\subsection{Experiment Result}

\begin{table*} [h]
    
    \centering
    \begin{threeparttable}
    \begin{tabular*}{0.83\linewidth}{l|c|c|c|c|c|c|c}
    \toprule [1.5pt]
    \multirow{2}*{Methods} & \multirow{2}*{Task} & \multicolumn{4}{c|}{SeqAcc} &  \multirow{2}*{FID} & \multirow{2}*{LPIPS}\\
    \cline{3-6} 

    & & ICDAR13(8ch) & ICDAR13 & TextSeg & LAION-OCR & &  \\
    \midrule [1.5pt]
    AnyText  & \multirow{4}*{Recon} & 0.89 & 0.87 & 0.81 & 0.86 & 22.73 & 0.0651   \\
    UDiffText &   & 0.94 & 0.91 & 0.93 & 0.90 & 15.79 & 0.0564   \\
    DreamText  &  & 0.95 & 0.94 & 0.96 & 0.93 & 12.13 & \textbf{0.0328}   \\
    UM-Text  &  & \textbf{0.99} &\textbf{ 0.98 }& \textbf{0.97} & \textbf{0.96} & \textbf{6.57} & 0.0479   \\
    \midrule [1.5pt]
    AnyText & \multirow{4}*{Editing} & 0.81 & 0.79 & 0.80 & 0.72 & - & -   \\
    UDiffText  &  & 0.84 & 0.83 & 0.84 & 0.78 & - & -   \\
    DreamText &  & 0.87 & 0.89 & 0.91 & 0.88 & - & -   \\
    UM-Text  &  & \textbf{0.93} & \textbf{0.93} & \textbf{0.95} & \textbf{0.93} & - & -   \\
    \bottomrule [1.5pt]
    \end{tabular*}
    \end{threeparttable}
    \caption{Comparison on the UDiffText benchmark dataset: The Recon task involves reconstructing text from the original image, while the Editing task focuses on modifying the text within the image.}\label{tab:UDiffText}
\end{table*}

\subsubsection{Quantitative Results}
We comprehensively evaluate UM-Text and state-of-the-art methods using the AnyText-benchmark, UDiffText benchmark, and our self-constructed UMT-benchmark. As shown in Table \ref{tab:anytext}, on the AnyText-benchmark, our method consistently outperforms competing approaches for both Chinese and English text across all metrics, including OCR accuracy (Sen.ACC, NED) and realism (FID, LPIPS). As shown in Table \ref{tab:UDiffText}, our method outperforms previous approaches in SeqAcc and FID, although our LPIPS score is lower than DreamText’s. This may be because our method produces colors and textures that better match the image style during text reconstruction.

\begin{table} [h]
    \label{tab_dwsc1}
    \centering

    \begin{tabular*}{0.99\linewidth}{l|c|c|c|c}
    \toprule [1.5pt]
    \multirow{2}*{Methods} & \multicolumn{2}{c|}{English} & \multicolumn{2}{c}{Chinese}  \\
    \cline{2-3} \cline{4-5}
    & Sen.ACC& NED&Sen.ACC& NED\\
    \midrule [1.5pt]
    Flux-Kontext & 0.325 & 0.502 & - & - \\ 
    Step1X-Edit & 0.358 & 0.524 & - & - \\ 
    OmniGen2 & 0.371 & 0.541 & - & - \\ 
    AnyText & 0.518 & 0.643 & 0.557 & 0.706 \\ 
    AnyText-2 & 0.693 & 0.723 & 0.720 & 0.806 \\ 
    UM-Text &  \textbf{0.790 }& \textbf{0.866} & \textbf{0.956} & \textbf{0.981} \\ 
    \bottomrule [1.5pt]
    \end{tabular*}
    \caption{Comparison on UMT-benchmark. Please note that all methods use the layout and text by UM-Designer.}
    \label{tab:UMT-benchmark}

\end{table}

Our UM-Designer model is capable of designing both layout and text, which motivates us to propose the UMT-benchmark to evaluate the performance of the entire pipeline. The UM-Designer model can also be integrated with other text editing models, such as AnyText and AnyText2, to generate product posters from clean product images. For fair comparison, our generative model, like previous state-of-the-art methods, is trained on the AnyWord-3M dataset without using any product-specific data. As shown in Table \ref{tab:UMT-benchmark}, we compare the Sen.ACC and NED metrics for both Chinese and English, and our method significantly outperforms previous approaches on both metrics.
\begin{figure}[h]
    \centering
    \includegraphics[width=0.47\textwidth]{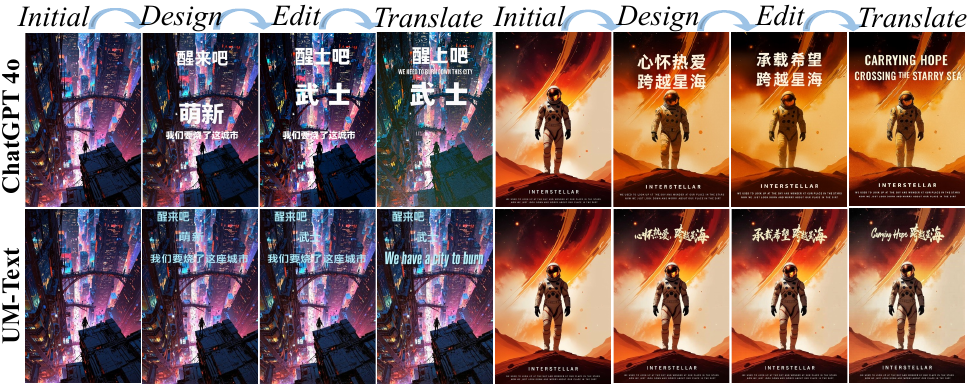}
    \caption{Compare UM-Text and ChatGPT4o in multi-turn image editing using natural language instructions, specifically in poster design, image editing, and image translation.}
    \label{fig:compare_4o}
\end{figure}
\subsubsection{Qualitative Results}
We conduct qualitative comparisons with state-of-the-art methods, including AnyText, AnyText-2, and FLUX-Text, on both English and Chinese multi-line text scenarios, as illustrated in Fig.\ref{fig:comppare_1}. Our method demonstrates superior performance in generating accurate, coherent, and visually harmonious text that blends seamlessly with the background, under complex conditions for both languages. In contrast, AnyText and AnyText-2 frequently produce results with blurred characters, duplicated text, or even incorrect glyphs. FLUX-Text generates text that is visually inconsistent with the background, suffers from color distortion, and also exhibits glyph errors, particularly in complex Chinese text scenarios. Notably, our method maintains precise glyph integrity and strong background consistency, even in challenging multi-line text settings. Meanwhile, we also conducte a multi-turn task comparison with ChatGPT-4o, as shown in Fig \ref{fig:compare_4o}. Our method maintains precise glyph integrity and consistency with the background, even in complex multi-line scenarios, whereas ChatGPT-4o often introduces unnecessary text modifications.

\subsection{Ablation Study}
We randomly sampled 100k images from the AnyWord-3M dataset, including 50k Chinese and 50k English images. To evaluate the contribution of each module in our method, we conducte ablation studies on the AnyText-benchmark by training for 10 epochs on this small-scale dataset, as shown in Table \ref{tab: Ablation study}. We used FLUX-Fill as the baseline, which demonstrates a lack of Chinese text generation capability. In addition, we compared the effect of adding a character-level visual encoder, which significantly improved the text generation ability of the baseline. We verified that the VLM embedding further enhanced the accuracy of text generation. Finally, we evaluated the impact of $\mathcal{L}_{RCL}$ and $\mathcal{L}_{RCI}$, which obtained an improvement of $4.8\%$ and $4.2\%$ respectively.

\begin{table} [h]
    \label{tab_dwsc}
    \centering
    \begin{tabular*}{0.95\linewidth}{l|c|c|c|c}
    \toprule [1.5pt]
    \multirow{2}*{Methods} & \multicolumn{2}{c|}{English} & \multicolumn{2}{c}{Chinese}  \\
    \cline{2-3} \cline{4-5}
    & Sen.ACC& NED& Sen.ACC& NED\\
    \midrule [1.5pt]
    Baseline & 0.309 & 0.469 & 0.029 & 0.062\\ 
    +Visual & 0.759 & 0.887 &  0.676 & 0.839\\ 
    +VLM & 0.782 & 0.901 &  0.698 & 0.848\\ 
    +RCL Loss & 0.799 & 0.915 &  0.725 & 0.856\\ 
    +RCI Loss & \textbf{0.824} & \textbf{0.925}  &  \textbf{0.746} & \textbf{0.863}\\ 
    \bottomrule [1.5pt]
    \end{tabular*}
    \caption{Ablation experiments of UM-Text conducted on a subset of the AnyWord-3M dataset.}
    \label{tab: Ablation study}
\end{table}

\section{Conclusion}
In this paper, we introduce UM-Text, a novel unified multimodal method designed to accomplish complex visual text editing tasks via simple natural language instructions. We explore a three-stage joint training strategy that integrates VLM and diffusion models, and propose the UM-Designer module for layout and text planning. Furthermore, we present the UM-Encoder, which fuses VLM embeddings, character-level visual embeddings, and T5 embeddings to enhance the model’s understanding of both scene images and text glyphs, thereby enabling accurate and style-consistent editing and generation of textual and visual content. To supervise fine-grained visual text glyph information, we propose regional consistency loss. In addition, we contribute UM-DATA-200K, a large-scale and diverse dataset of layouts and texts, as well as the UMT-benchmark for evaluating instruction-based visual text editing. Extensive qualitative and quantitative results demonstrate the superiority of our approach.

\bibliography{aaai2026}
\end{document}